\documentclass[10pt,twocolumn,letterpaper]{article}

\usepackage[pagenumbers]{cvpr} 

\usepackage{graphicx}
\usepackage{amsmath}
\usepackage{amssymb}
\usepackage{booktabs}
\usepackage{multirow}
\usepackage{colortbl}
\definecolor{LightCyan}{rgb}{0.88,1,1}
\usepackage[accsupp]{axessibility}

\usepackage[pagebackref,breaklinks,colorlinks]{hyperref}

\usepackage[capitalize]{cleveref}
\crefname{section}{Sec.}{Secs.}
\Crefname{section}{Section}{Sections}
\Crefname{table}{Table}{Tables}
\crefname{table}{Tab.}{Tabs.}

\def\cvprPaperID{40} 
\def\confName{CVPR}
\def\confYear{2023}

\begin{document}

\title{Zoom-VQA: Patches, Frames and Clips Integration for \\ 
Video Quality Assessment}

\author{Kai Zhao, ~Kun Yuan, ~Ming Sun and Xing Wen \\
Kuaishou Technology \\
{\tt\small \{zhaokai05,yuankun03,sunming03,wenxing\}@kuaishou.com}
}
\maketitle

\begin{abstract}

Video quality assessment (VQA) aims to simulate the human perception of video quality, which is influenced by factors ranging from low-level color and texture details to high-level semantic content. To effectively model these complicated quality-related factors, in this paper, we decompose video into three levels (\ie, patch level, frame level, and clip level), and propose a novel Zoom-VQA architecture to perceive spatio-temporal features at different levels. It integrates three components: patch attention module, frame pyramid alignment, and clip ensemble strategy, respectively for capturing region-of-interest in the spatial dimension, multi-level information at different feature levels, and distortions distributed over the temporal dimension. Owing to the comprehensive design, Zoom-VQA obtains state-of-the-art results on four VQA benchmarks and achieves 2nd place in the NTIRE 2023 VQA challenge. Notably, Zoom-VQA has outperformed the previous best results on two subsets of LSVQ, achieving 0.8860 (+1.0\%) and 0.7985 (+1.9\%) of SRCC on the respective subsets. Adequate ablation studies further verify the effectiveness of each component. 
Codes and models are released in \url{https://github.com/k-zha14/Zoom-VQA}.

\end{abstract}

\section{Introduction}
\label{sec:intro}


To achieve the ultimate viewing experience, streaming platforms often employ diverse enhancement techniques to ameliorate the quality of uploaded content \cite{DBLP:journals/tog/XiaoKFCL18, DBLP:journals/tog/XiaoNCFLK20, DBLP:conf/cvpr/HeSCFD22}. The primary phase in this endeavor involves precise evaluation of the perceptual quality of the content, which can be achieved by subjective and objective quality assessment measures. Given that subjective quality assessment experiments can be time-consuming, expensive, and complex \cite{DBLP:conf/cvpr/CheonYKL21, DBLP:conf/cvpr/GuoBHZL21}, both industry and academia are engaged in the pursuit of establishing more scalable and feasible objective quality assessment methods, capable of managing ever-growing volumes of user-generated content \cite{gao2023vdpve}.

Generally, VQA can be classified into three categories: full-reference, reduced reference, and no-reference, based on the availability of reference information \cite{DBLP:conf/mm/MaF21}. Despite evident differences in the reference feature fusion, almost all VQA methods typically follow a conventional paradigm \cite{DBLP:journals/tip/BosseMMWS18}, that is, extracting visual features related to perceived quality and designing regression or classification head for quality prediction from the extracted features. 

\begin{figure}[t]
  \centering
  \includegraphics[width=0.96\linewidth]{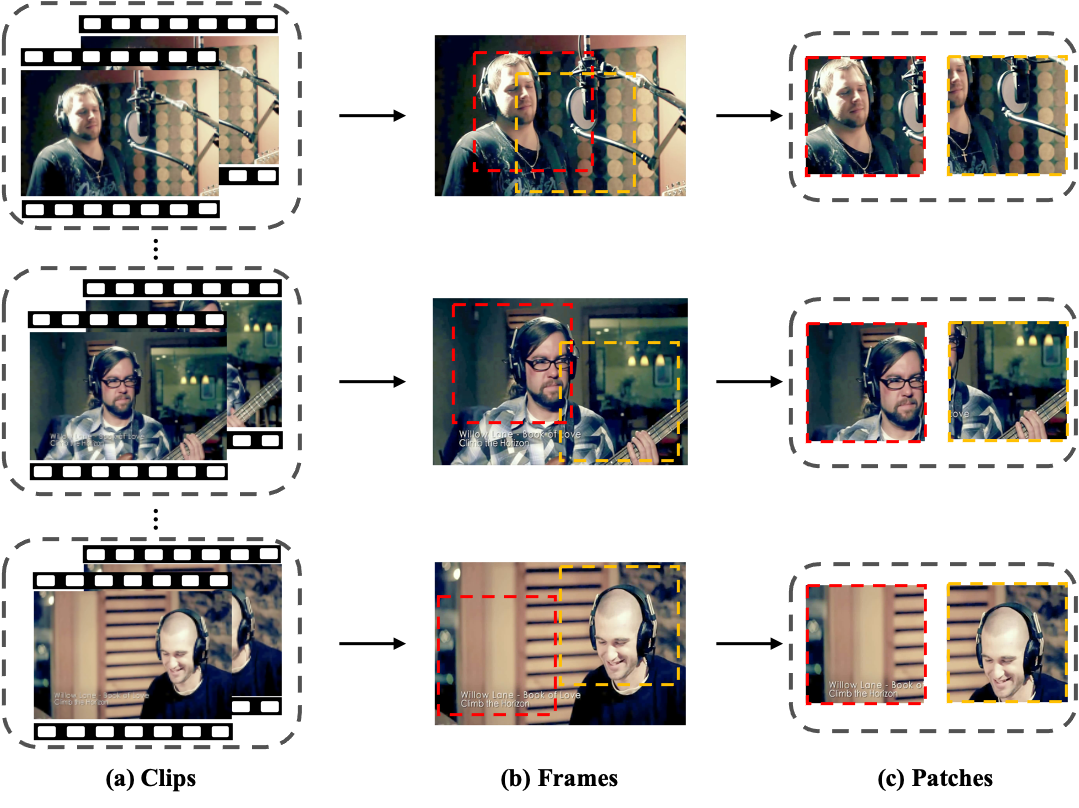}
  \caption{The decomposition of a typical video into three levels, (a) clips, (b) frames, and (c) patches. Videos captured at different levels contain various factors that affect video quality. For instance, motion blur is reflected in the clip level, semantic information is contained in the frame level, and local block artifacts and noise are manifested in the patch level.}
  \label{fig:deomcpose}
  \vspace{-0.5cm}
\end{figure}

Established on this paradigm, the success of VQA methods hinges on two aspects: 1) extracting more powerful and representative features, including advanced backbones and modules \cite{DBLP:journals/tcsv/LiZTZW22, DBLP:conf/eccv/WuCHLWSYL22}, quality-aware pre-training \cite{DBLP:journals/tip/MadhusudanaBWAB22, DBLP:conf/cvpr/QPT} and multiple hand-crafted features ensemble \cite{DBLP:journals/corr/GhadiyaramB16, DBLP:journals/tip/TuWBAB21}, and 2) designing quality score prediction head, such as normal regression heads \cite{DBLP:conf/mm/SunMLZ22, DBLP:journals/tip/TuWBAB21}, patches fusion \cite{DBLP:conf/cvpr/YingMGB21} and IP-NLR \cite{DBLP:conf/eccv/WuCHLWSYL22}. Additionally, some studies explore patches \cite{DBLP:conf/cvpr/YingMGB21}, frames \cite{DBLP:conf/mm/SunMLZ22} and clip integration \cite{DBLP:conf/eccv/WuCHLWSYL22} separately, inspired by locality and sensitivity of human visual system (HVS). Along with state-of-the-art deep neural networks \cite{DBLP:conf/cvpr/0003MWFDX22, DBLP:conf/cvpr/LiuN0W00022}, we take a step further by integrating patches, frames and clips together. We claim that the performance of VQA methods could be further improved through this method.

Specifically, we treat a video from a \textit{zooming in} perspective. As shown in \cref{fig:deomcpose}, a typical video can be decomposed into clips, frames and patches progressively. However, due to the absence of fine-grained labels (\eg clip-level, frame-level and patch-level) in existing VQA datasets \cite{DBLP:conf/qomex/HosuHJLMSLS17, DBLP:conf/mmsp/WangIA19}, the multi-level supervised training and integration is practically difficult. Besides, with the popularization of deep learning-based video enhancement methods \cite{DBLP:conf/iccvw/WangXDS21, DBLP:journals/ijon/LiYCCFXLC22}, nonexistent artifacts are generated to improve perceptual aspect of processed videos. Consequently, the precise prediction of perceived quality has become even more challenging.

To overcome the above difficulties, we propose Zoom-VQA, a multi-level integration framework for VQA. The contributions of this framework are summarized as follows:
\begin{itemize}
    \item To capture region-of-interest in the spatial dimension that affects quality, a \textit{patch attention module} is proposed that generates weights and scores for patches.
    \item To obtain distortions that are reflected in different feature levels, a \textit{frame pyramid alignment} module is proposed to integrate multi-scale information.
    \item To model spatio-temporal information effectively, a \textit{clip ensemble strategy} is utilized to aggregate temporal information adaptively.
    \item Zoom-VQA obtains state-of-the-art results on four VQA benchmarks and achieves \textbf{2nd place} in the NTIRE 2023 VQA challenge. Notably, Zoom-VQA has outperformed the previous best results on two subsets of LSVQ, achieving \textbf{0.8860} (+1.0\%) and \textbf{0.7985} (+1.9\%) of SRCC on the respective subsets.
\end{itemize}


\section{Related work}
\label{sec:related}

\subsection{Video Quality Assessment}
The fundamental goal of objective VQA is to accurately predict the perceived quality of human viewers. To this end, a naive solution is to apply classic image quality assessment metrics \cite{DBLP:journals/tip/MittalMB12, DBLP:journals/spl/MittalSB13, DBLP:conf/cvpr/YeKKD12} on each video frame, and then perform a feature/score pooling stage. Since the efficacy of temporal pooling is temporal-dependent \cite{DBLP:conf/icip/TuCCBAB20}, it is non-trivial to incorporate spatio-temporal information. V-BLINDS \cite{DBLP:journals/tip/SaadBC14} proposes a spatio-temporal natural scene statistics (NSS) model by quantifying the frame difference of NSS features. VIIDEO \cite{DBLP:journals/tip/MittalSB16} exploits the inherent statistical  characteristics of natural videos to deal with distortion-specific disturbances. TLVQM \cite{DBLP:journals/tip/Korhonen19} and VIDEVAL \cite{DBLP:journals/tip/TuWBAB21} combine abundant spatio-temporal features and typical quality assessment methods to regress them into the video quality score. However, these hand-crafted features are not sufficient to capture complicated factors affecting video quality, such as semantic information. Subsequent works \cite{DBLP:conf/mm/KorhonenSY20, DBLP:journals/corr/abs-2101-10955} attempt to integrate hand-crafted features with semantic features extracted by shallow pre-trained CNN models. 

Recently, deep learning-based VQA methods have become dominant in the VQA field, with the rise of large-scale VQA datasets \cite{DBLP:conf/qomex/HosuHJLMSLS17, DBLP:conf/mmsp/WangIA19, DBLP:conf/cvpr/YingMGB21}. Among them, VSFA \cite{DBLP:conf/mm/LiJJ19} utilizes the pre-trained ResNet-50 \cite{DBLP:conf/cvpr/HeZRS16} from ImageNet-1k \cite{DBLP:conf/cvpr/DengDSLL009} to extract spatial features, and then models the temporal relationship by GRU \cite{DBLP:conf/emnlp/ChoMGBBSB14}. Based on this, MDVSFA \cite{DBLP:journals/ijcv/LiJJ21} explores a mixed dataset training strategy with VSFA to mitigate the cross-dataset evaluation challenge. MLSP-FF \cite{DBLP:journals/access/Gotz-HahnHLS21} adopts deeper and heavier Inception-ResNet-V2 for feature extraction. Bi-LSTM \cite{DBLP:journals/tsp/SchusterP97} and graph convolution netowrk \cite{DBLP:conf/iclr/KipfW17} are also introduced into the VQA field, to enhance temporal feature extraction \cite{DBLP:conf/mm/XuLZZW021}. In addition, some methods \cite{DBLP:conf/cvpr/YingMGB21, DBLP:journals/tcsv/LiZTZW22, DBLP:journals/corr/abs-2206-09853} use video models (\eg 3D-CNN \cite{DBLP:conf/cvpr/CarreiraZ17, DBLP:journals/corr/abs-1812-03982}) pre-trained on action recognition datasets to extract spatial-temporal features. However, limited by high computational cost on high-resolution videos, the above methods usually regress the subjective quality scores with fixed features instead of end-to-end training \cite{DBLP:journals/tcsv/LiZTZW22}. Aimed at this issue, FAST-VQA \cite{DBLP:conf/eccv/WuCHLWSYL22} and DOVER \cite{DBLP:journals/corr/abs-2211-04894} propose grid mini-patch sampling (GMS) strategy, which samples patches at their raw resolution to maintain the global quality with the original video. In order to accommodate high-resolution inputs of the challenge dataset, implementation and modification of GMS will be further discussed in \cref{sec:cn}. 

\subsection{Advancement in Visual Networks}
The advancement of visual backbone networks prospers various downstream tasks tremendously \cite{DBLP:conf/iccvw/LiangCSZGT21, DBLP:conf/cvpr/LiuN0W00022, DBLP:journals/corr/abs-2111-09886, DBLP:conf/eccv/CarionMSUKZ20,DBLP:conf/iccv/YuanG0ZYW21}. Based on the modality of input data, visual networks can be classified into two types, image networks (\ie 2D networks) and video networks (\ie 3D networks). C3D \cite{DBLP:conf/iccv/TranBFTP15} pioneeringly devises an 11-layer CNN with 3D-CNN to adapt to video inputs. Subsequent P3D \cite{DBLP:conf/iccv/QiuYM17}, S3D \cite{DBLP:conf/eccv/XieSHTM18} and R(2+1)D \cite{DBLP:conf/cvpr/TranWTRLP18} observe that disentangled spatial and temporal convolutions results in a more favorable speed-accuracy trade-off than the pure 3D convolution. Remarkably, I3D \cite{DBLP:conf/cvpr/CarreiraZ17} reveals that successful 2D networks could be seamlessly \textit{inflated} to corresponding 3D networks and even their parameter. 

In recent days, a shift in backbone architecture, from CNNs to Transformers (ViT) \cite{DBLP:conf/iclr/DosovitskiyB0WZ21, DBLP:conf/icml/TouvronCDMSJ21}, has begun. Especially, Swin Transformer \cite{DBLP:conf/iccv/LiuL00W0LG21} reintroduces the inductive bias of convolutions (\ie, locality, translation invariance and hierarchy), which enables it to serve as a general-purpose backbone. The success of image Transformer leads to further investigation of Transformer-based video networks (\eg, ViViT \cite{DBLP:conf/iccv/Arnab0H0LS21}, MViT \cite{DBLP:conf/iccv/0001XMLYMF21}, Video Swin Transformer \cite{DBLP:conf/cvpr/LiuN0W00022}). Among all characteristics of Transformers, the patch-wise operations inherently differentiate the edges of patches, thus making them ideal for handling input sampled by GMS.


\begin{figure*}[t]
  \centering
    \includegraphics[width=1.0\linewidth]{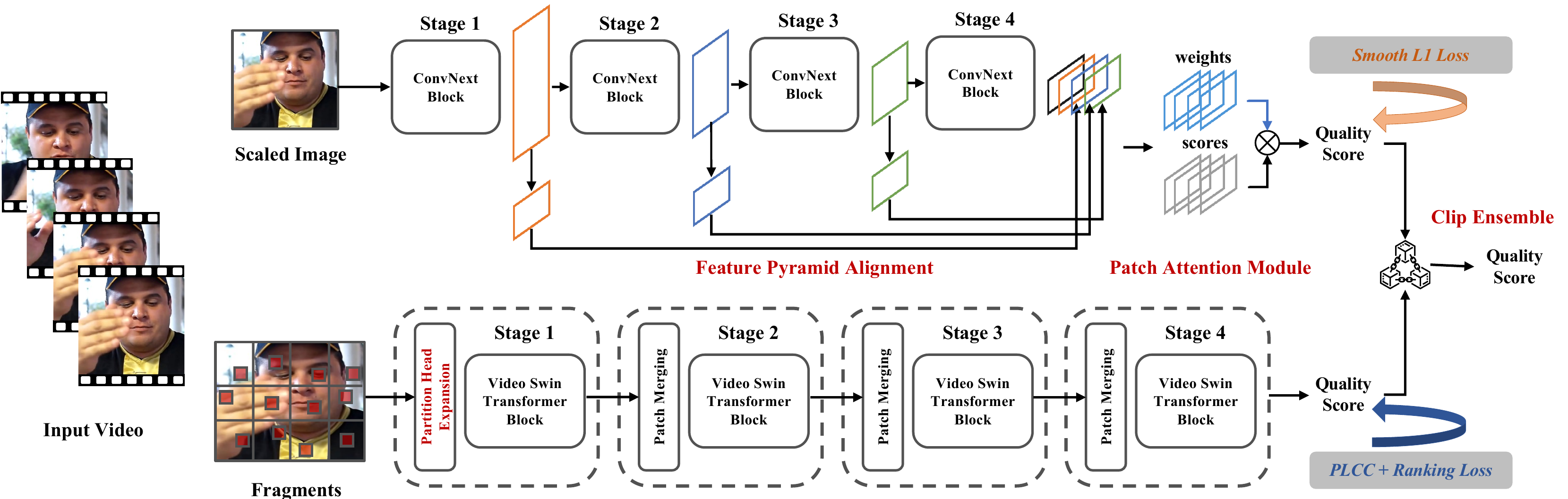}
  \caption{Illustration of the framework of Zoom-VQA. The overall architecture consists of two parts. One part is a perception network based on image input that obtains global information. The other part is a perception network based on video clip input, obtaining texture information by sampling fragment regions. During training, the two parts are trained separately using different optimization objectives. During inference, we use an ensemble to combine the prediction results of the two parts to enhance the model's generalization ability.}
  \label{fig:framework}
\end{figure*}

\section{Method}
\label{sec:method}


We propose a novel Zoom-VQA network architecture that evaluates video quality by comprehensively analyzing the ``patch-frame-clip" dimensions of the video.
Through this approach, Zoom-VQA can effectively capture both local and global information that impacts video quality, making it more suitable for assessing enhanced videos.
In this section, we first give the overall framework in \cref{sec:framework} and adopt a perspective that gradually "zooms out" from the micro to the macro level to introduce Zoom-VQA, covering a \textit{patch attention module} which considers qualities of different spatial regions in \cref{sec:pam}, a \textit{frame pyramid alignment module} which tackles multi-level feature information in \cref{sec:fpa} and a \textit{clip ensemble strategy} which utilizes the spatio-temporal information fully in \cref{sec:cn}. Last, the optimization objective will be given in \cref{sec:loss}.

\subsection{Overall Framework}\label{sec:framework}

As given in \cref{fig:framework}, the overall framework of Zoom-VQA consists of two parts. Given an input video $\mathcal{V}$, one part is a perception network based on image input (named IQA branch) $\mathbf{x}_t$, which denotes the $t$-th frame sampled from $\mathcal{V}$, that obtains global information using ConvNext \cite{DBLP:conf/cvpr/0003MWFDX22} as the backbone, analyzes the video frame by frame, and outputs the average results as the final quality prediction score. The other part is a perception network based on video segment input (named VQA branch) $\mathcal{G}=\{\mathbf{g}_t, \cdots, \mathbf{g}_T\}$ (will be described below), using Video Swin Transformer \cite{DBLP:conf/cvpr/LiuN0W00022} as the backbone, and obtaining  spatio-temporal information by sampling fragment regions as FAST-VQA \cite{DBLP:conf/eccv/WuCHLWSYL22}, outputting the overall quality prediction score. The final prediction of the quality score $y^{\prime}$ is obtained through an ensemble strategy, enhancing the model's generalization ability.

\subsection{Patch Attention Module}\label{sec:pam}


To comprehensively consider the quality of different regions in the spatial dimension, we added a patch-based attention module to the head of the IQA branch. Specifically, as given in \cref{fig:framework}, this module receives extracted features as input and generates \textit{weights} and \textit{scores} through two separate branches consisting of lightweight networks. The quality score of each patch in the feature map can be obtained by multiplication of corresponding score and weight in a self-attention manner. Then the final prediction of the whole image is generated by summarizing all patches. Given the input features $\mathbf{f}_t\in \mathbb{R}^{c\times h \times w}$, this can be noted as follows:
\begin{equation}\label{eq:pam}
    \begin{aligned}
        \mathbf{w}_{\text{patch}} & = \text{ReLU}(\text{FC}(\text{ReLU}(\text{FC}(\mathbf{f}_t)))), \\
        \mathbf{s}_{\text{patch}} & = \text{Sigmoid}(\text{FC}(\text{ReLU}(\text{FC}(\mathbf{f}_t)))), \\
        y_t^{\text{IQA}} & = \sum \mathbf{w}_{\text{patch}} \odot \mathbf{s}_{\text{patch}},
    \end{aligned}
\end{equation}
where $\mathbf{w}_{\text{patch}}\in \mathbb{R}^{c\times h \times w}$ is the learned weights for each patch, and $\mathbf{s}_{\text{patch}}\in \mathbb{R}^{c\times h \times w}$ is corresponding scores. The $\odot$ indicates the Hadamard product and $y_t^{\text{IQA}}$ is the prediction score for the $t$-th frame. 
By this explicit constraint, the network can pay more attention to  region-of-interest in the frame, capturing features about distortions that are non-uniformly distributed across spatial dimensions.

\subsection{Frame Pyramid Alignment}\label{sec:fpa}


Video quality is determined by multiple aspects, including content, distortions, and compression artifacts \cite{DBLP:conf/mmsp/WangIA19}. The factors that may affect quality have different responses to features at different scales. For example, content-related semantic information is usually reflected in high-level features, while distortion-related information such as texture and color is more reflected in low-level features. To obtain these multi-level features, we design a frame pyramid alignment module in the IQA branch, performing a bottom-up pathway. As given in \cref{fig:framework}, features at different stages are extracted, resulting in $\mathbf{f}_l\in \mathbb{R}^{c_l\times h_l \times w_l}$ for the $l$-th stage. Then, an adaptive average pooling operation is performed on the spatial dimension to align the spatial scale with the last stage of features. Lastly, the aligned features from different stages are concatenated and sent into the patch attention module. Given the ConvNext-Tiny with 4 stages, this procedure can be noted as follows:
\begin{equation}
    \begin{aligned}
        \mathbf{f}^{\prime}_l = & \text{AdaptiveAvgPool}(\mathbf{f}_l, {h_l}/{h_4})|_{l=1}^3, \\
        & \text{where} ~ \mathbf{f}^{\prime}_l\in \mathbb{R}^{c_l\times h_4 \times w_4}, \\
        \mathbf{f}^{\prime}_t = &\text{Concatenate}\{\mathbf{f}^{\prime}_1, \mathbf{f}^{\prime}_2, \mathbf{f}^{\prime}_3, \mathbf{f}^{\prime}_4\}, \\ 
        & \text{where} ~ \mathbf{f}^{\prime}_t\in \mathbb{R}^{\sum_{l=1}^4 c_l \times h_4 \times w_4},
    \end{aligned}
\end{equation}
where ${h_l}/{h_4}$ is the pooling ratio. Then the input $\mathbf{f}_t$ for the patch attention module in \cref{eq:pam} can be replaced by $\mathbf{f}^{\prime}_t$ with more abundant features.

\begin{figure}[t]
  \centering
    \includegraphics[width=1.0\linewidth]{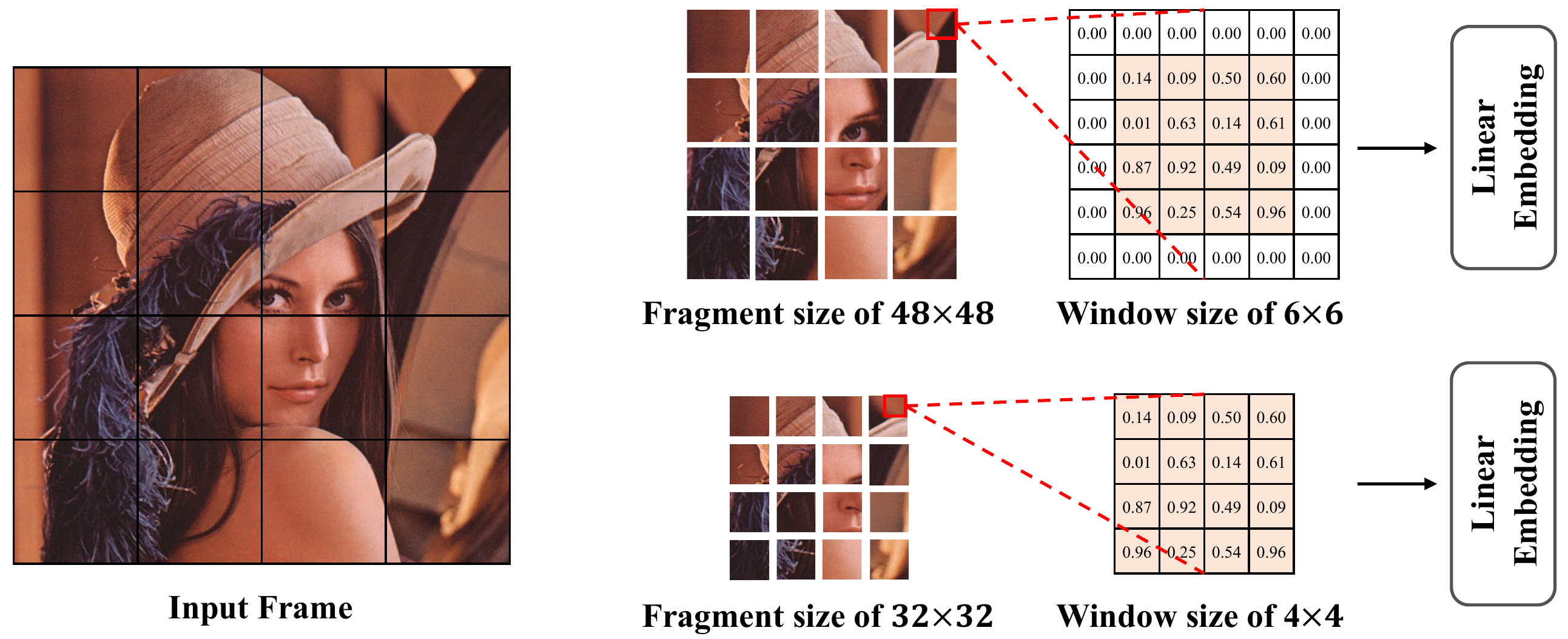}
  \caption{Patch head expansion for the larger size of fragments. The window size is expanded from 4 to 6 with zero padding during the weight initialization phase. In this type, the change in fragment size can be well-compatible with the pre-trained weights.}
  \label{fig:window}
\end{figure}

\subsection{Clip Ensemble Strategy}\label{sec:cn}


To reflect the impact caused by temporal jitters, we employed a clip ensemble strategy to attain the final prediction quality score of a video. For the IQA branch, the video-level prediction is obtained by averaging the predicted scores for all video frames, which is formulated as:
\begin{equation}
    y^{\text{IQA}} = \frac{1}{T}\sum_{t=1}^T y^{\text{IQA}}_t.
\end{equation}

For more effective spatio-temporal information extraction, we introduce a Video Swin Transformer based on video clip input in the VQA branch. To preserve the original video quality and obtain local texture information which benefits quality assessment, we utilize the sampling strategy of fragments used in FAST-VQA. The fragments are obtained through uniform grid mini-patch sampling. This method vastly reduces the computational cost by 97.6\% compared with computing attention on the whole resolution. It allows the network to pay more attention to low-quality texture information (\eg, noise, blur, blocky artifacts, etc.) and reduces its focus on higher-level semantics.


In practice, we found that appropriately increasing the size of the fragment helps to better model local information (\eg, expanding the size from 32 to 48), but changes in size can prevent the subsequent tokenization from utilizing existing pre-trained weights (\ie, mismatching of window size). To address this issue, we propose a patch head expansion module that performs zero-padding around existing convolutional kernels of tokenization as shown in \cref{fig:window}. This approach enables the pre-trained weights to be effectively utilized while increasing the receptive field for extracting low-level information. Then the expanded fragments are sent to the subsequent Video Swin Transformer, generating the quality prediction of $y^{\text{VQA}}$.

Finally, the prediction results of these two branches are aggregated to obtain the final video quality. To ensure consistency in the distribution of the predicted range, we use the sigmoid function for normalization. The clip ensemble strategy for the final quality prediction can be noted as:
\begin{equation}
    y^{\prime} = \frac{1}{2} \big(\text{Sigmoid}(y^{\text{IQA}})+\text{Sigmoid}(y^{\text{VQA}})\big).
\end{equation}

\subsection{Optimization Objective}\label{sec:loss}


During training, to ensure convergence, the two branches of Zoom-VQA are \textbf{trained separately using different optimization objectives}. For the IQA branch, the network is optimized using a smooth $\mathcal{L}_1$ regression loss. Given the labeled MOS $y$, the optimization objective can be noted as:
\begin{equation}
    \footnotesize
    \text{min}~\mathcal{L}_{reg} = \left \{
        \begin{array}{cc}
             0.5(y-y^{\text{IQA}})^2, & \text{if}~|y-y^{\text{IQA}}|<1 \\
             |y-y^{\text{IQA}}|-0.5,  & \text{otherwise.}
        \end{array}
    \right.
\end{equation}

For the VQA branch, a PLCC-induced loss and a ranking-based loss are utilized. Assume we have $m$ videos on the training batch. Given the predicted quality scores $\mathbf{Y}^{\text{VQA}} = \{y^{\text{VQA}}_1, y^{\text{VQA}}_2, \cdots, y^{\text{VQA}}_m\}$ and corresponding MOS values $\mathbf{Y} = \{y_1, y_2, \cdots, y_m\}$, the PLCC-induced loss is defined as:
\begin{equation}
    \footnotesize
    \mathcal{L}_{plcc} = 
    \Big(1-\frac{
    \sum_{i=1}^{m}(y^{\text{VQA}}_i - a^{\text{VQA}})(y_i-a)
    }{
    \sqrt{\sum_{i=1}^{m}(y^{\text{VQA}}_i - a^{\text{VQA}})^2\sum_{i=1}^{m}(y_i-a)^2}
    }\Big)/2,
\end{equation}
where $a^{\prime}$ and $a$ are the mean values of $\mathbf{Y}^{\text{VQA}}$ and $\mathbf{Y}$ respectively. And the ranking-based loss can be denoted as:
\begin{equation}
    \footnotesize
    \mathcal{L}_{rank} = \frac{1}{m^2} \sum_{i=1}^m\sum_{j=1}^m \text{max}(0,|y_i-y_j|\\-e(y_i,y_j)\cdot(y^{\text{VQA}}_i - y^{\text{VQA}}_j )),
\end{equation}
where $e(y_i,y_j)$ is $1$ if $y_i \ge y_j$, else is $-1$ if $y_i < y_j$. And the optimization objective can be written as: 
\begin{equation}
    \text{min}~ \mathcal{L}_{plcc} + \beta \cdot \mathcal{L}_{rank},
\end{equation}
where $\beta$ is the balancing coefficient which is set to 0.3.

\section{Experiments}
\label{sec:exp}

\subsection{Datasets and Evaluation Criteria}

\paragraph{Datasets.} 
\label{ssec:dataset}

\begin{figure}[t]
  \centering
  \begin{subfigure}{0.48\linewidth}
    \includegraphics[width=\columnwidth]{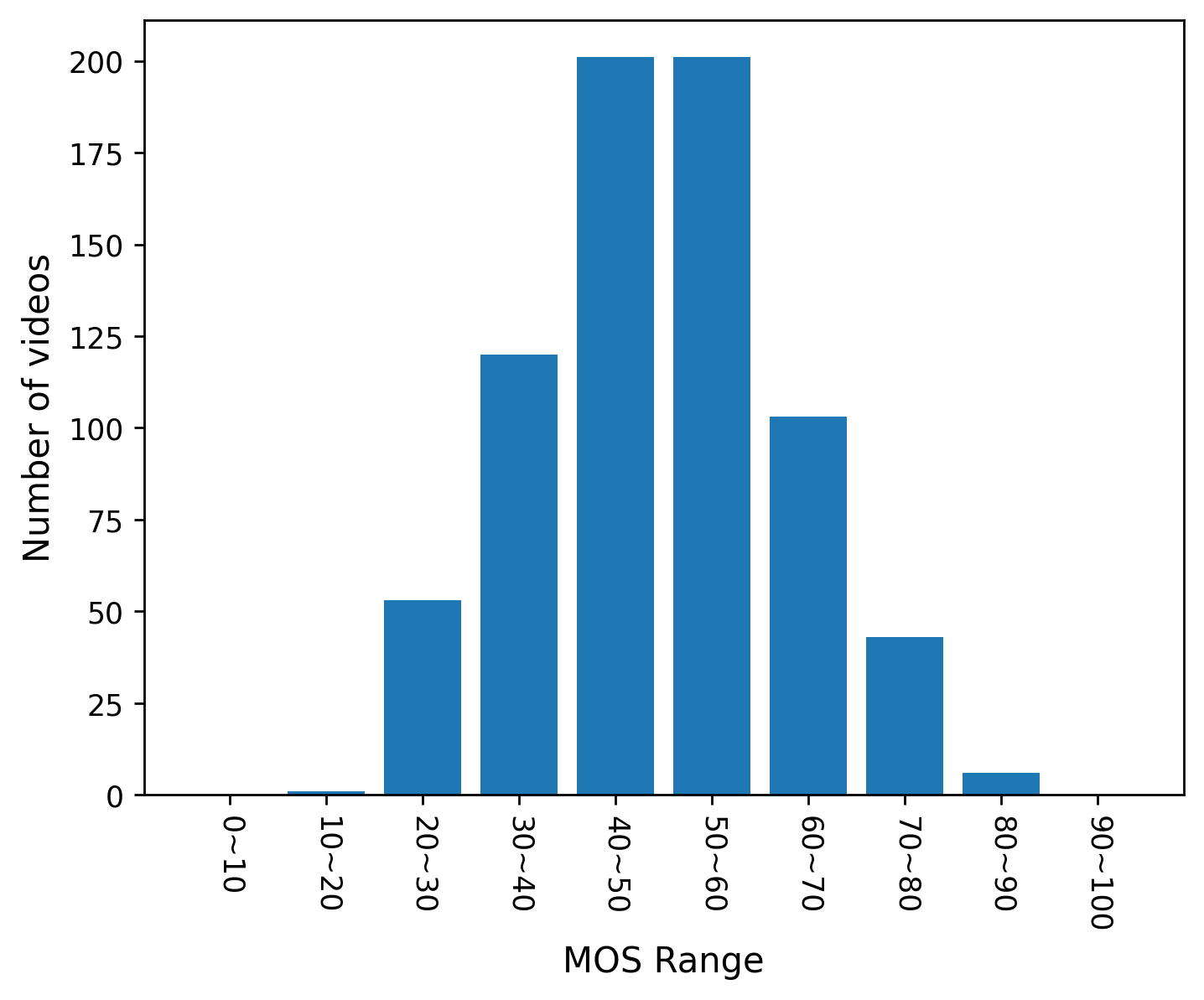}
    \caption{Training split}
  \end{subfigure}
  \begin{subfigure}{0.48\linewidth}
    \includegraphics[width=\columnwidth]{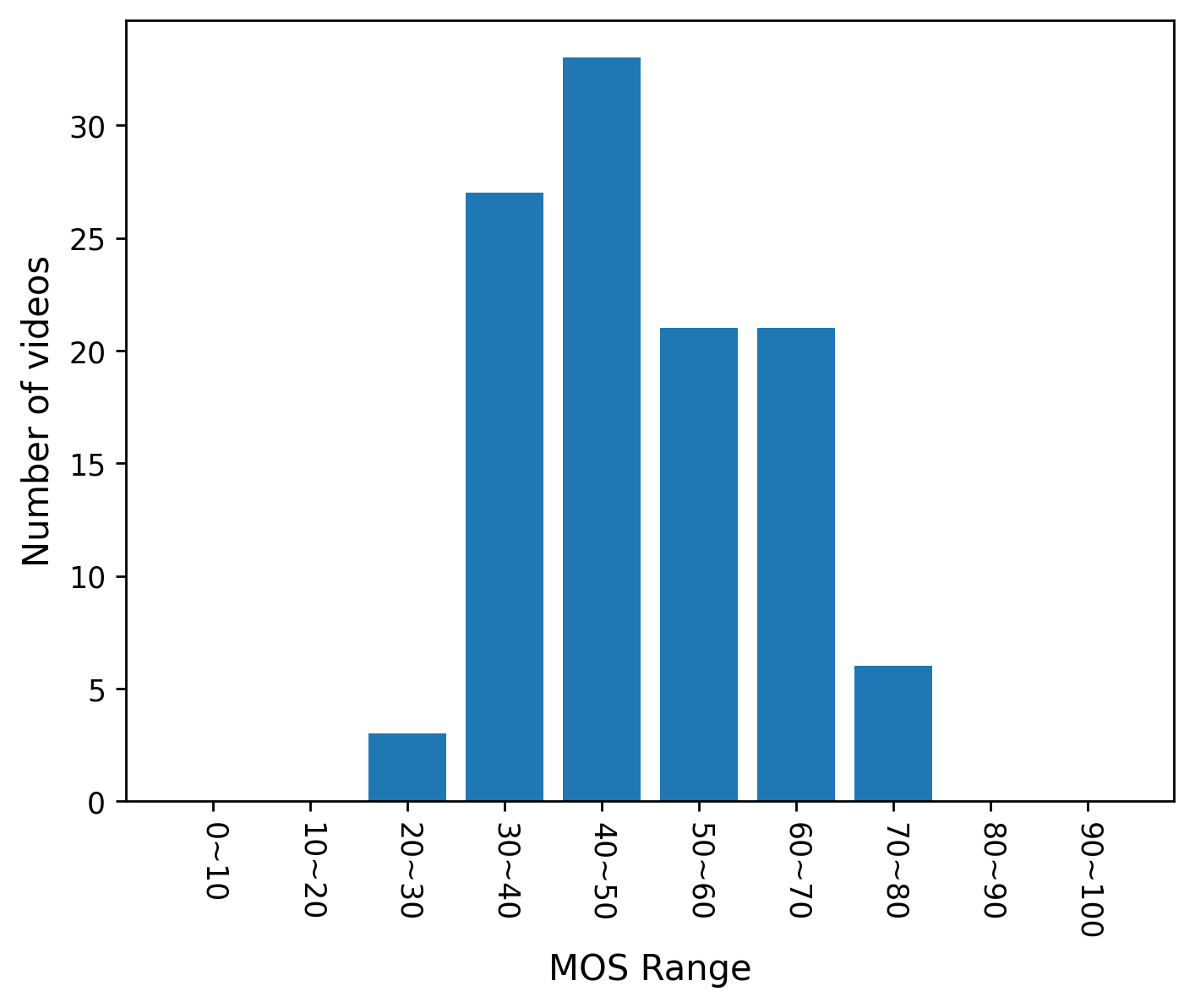}
    \caption{Testing split}
  \end{subfigure}
  \caption{Distribution of labeled MOS in the custom divided train-test splits of VDPVE.}
  \label{fig:mos}
  \vspace{-0.2cm}
\end{figure}

\begin{figure}[t]
  \centering
  \begin{subfigure}{0.48\linewidth}
    \includegraphics[width=\columnwidth]{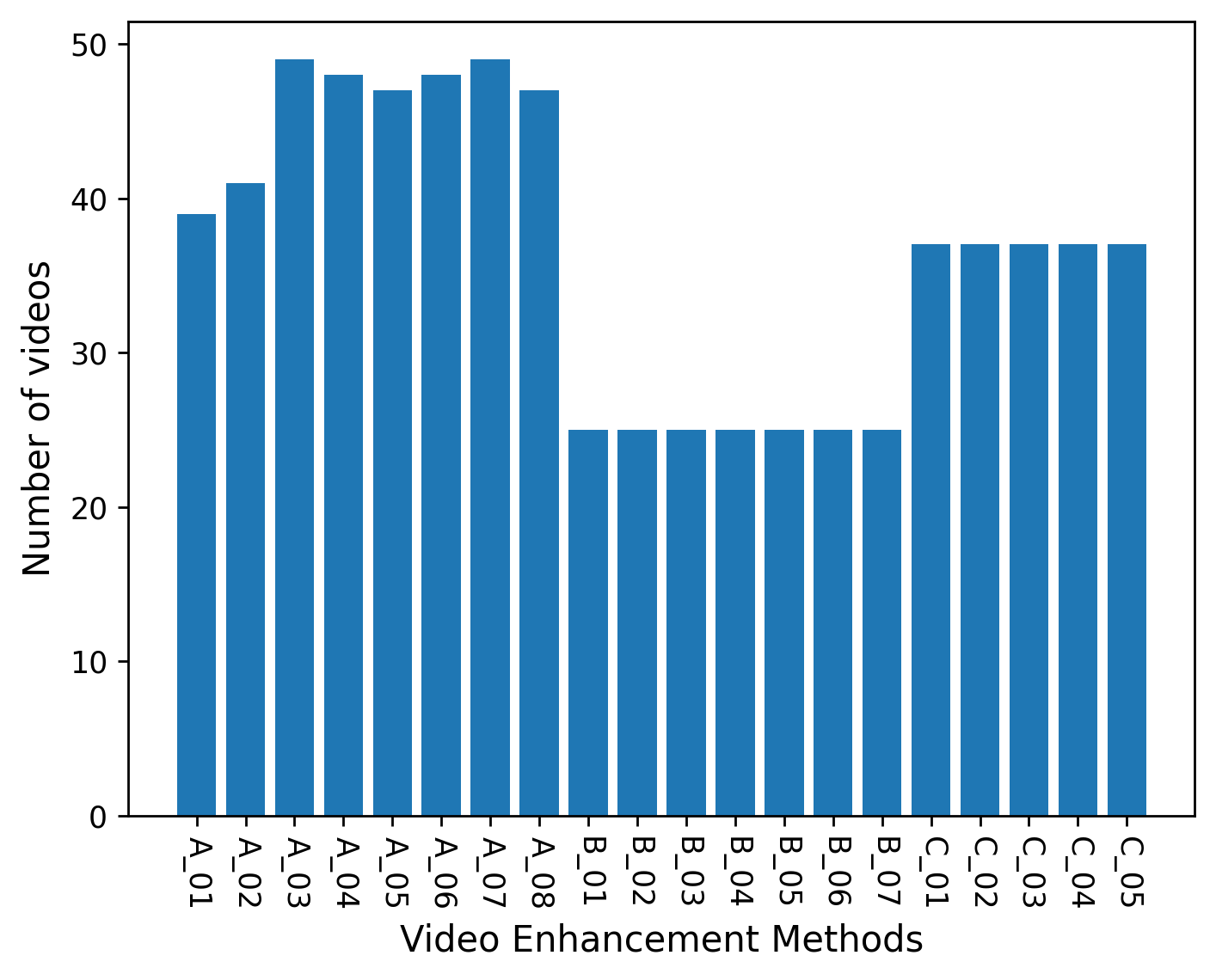}
    \caption{Training split}
  \end{subfigure}
  \begin{subfigure}{0.48\linewidth}
    \includegraphics[width=\columnwidth]{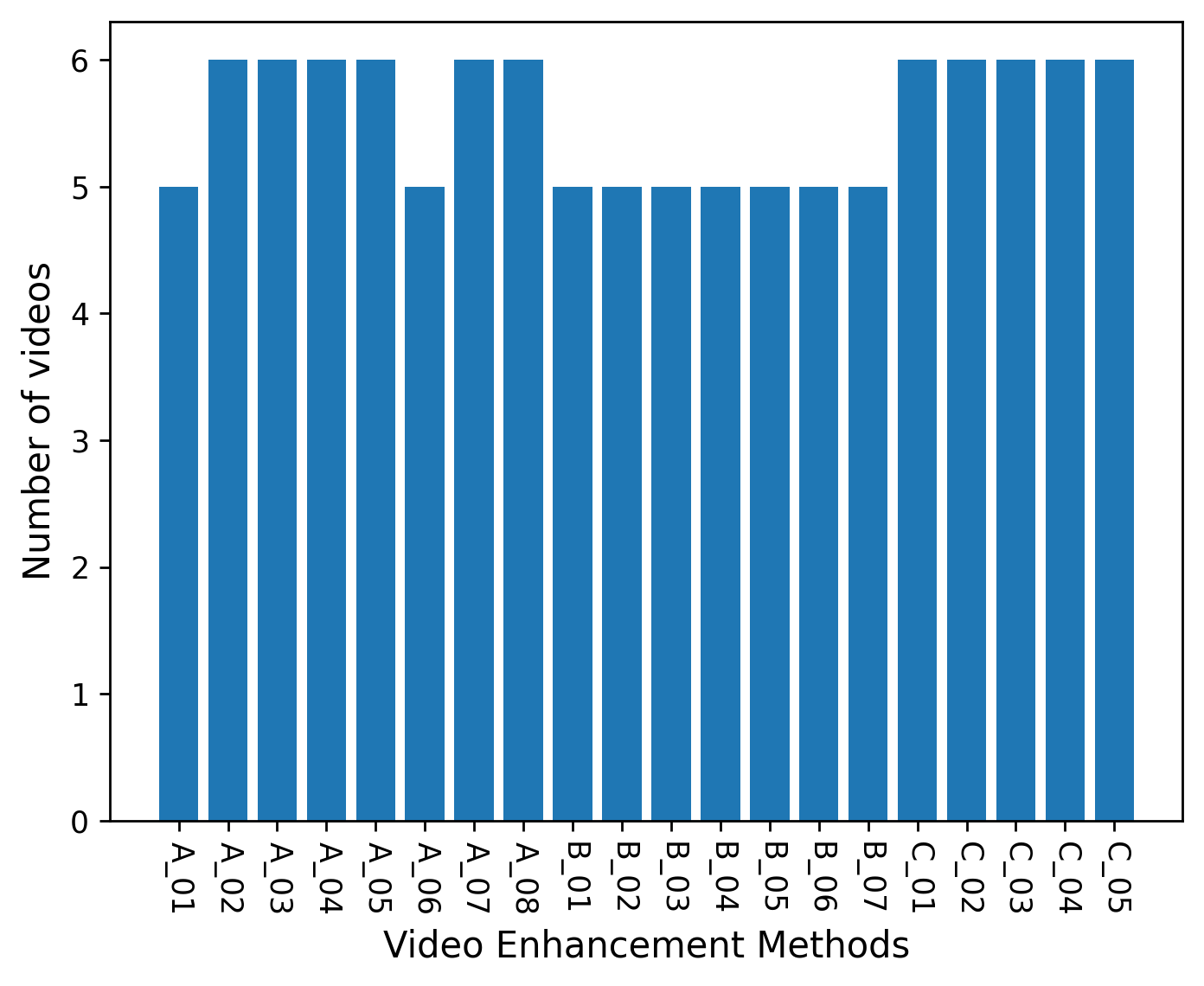}
    \caption{Testing split}
  \end{subfigure}
  \caption{Distribution of adopted video enhancement methods in the custom divided train-test splits of VDPVE.}
  \label{fig:methods}
  \vspace{-0.5cm}
\end{figure}

Four NR-VQA datasets are selected for evaluation, including VDPVE \cite{gao2023vdpve}, LSVQ \cite{DBLP:journals/tip/SinnoB19}, KoNViD-1k \cite{DBLP:conf/qomex/HosuHJLMSLS17} and LIVE-VQC \cite{DBLP:journals/tip/SinnoB19/live-vqc}. 
In detail, VDPVE has 1,211 videos with different enhancements, enabling the quality assessment of enhanced videos. Specifically, they can be divided into three sub-datasets. The first sub-dataset consists of 600 videos with color, brightness, and contrast enhancements. The second sub-dataset consists of 310 videos with deblurring. The third sub-dataset consists of 301 deshaked videos.
LSVQ, which is the largest VQA dataset by far, samples hundreds of thousands of open-source Internet UGC digital videos to match the feature distributions of social media UGC videos, collecting 39,000 real-world videos of diverse sizes, contents, and distortions.
LSVQ also provides two official test subsets, including LSVQ$_{\text{test}}$ and LSVQ$_{\text{1080p}}$. The LSVQ$_{\text{test}}$ consists of 7,400 various resolution videos from 240P to 720P, and LSVQ$_{\text{1080p}}$ consists of 3,600 1080P high resolution videos. KoNViD-1k contains 1,200 videos that are fairly filtered from a large public video dataset YFCC100M. The videos are 8 seconds long with 24/25/30 FPS and a resolution of $960 \times 540$. Besides, LIVE-VQC consists of 585 videos with complex authentic distortions captured by 80 different users using 101 different devices. 

In our experiments, VDPVE and LSVQ are used for training and evaluation. While KoNViD-1k and LIVE-VQC are only used for cross-dataset evaluation. The train-test split of LSVQ is aligned with previous studies \cite{DBLP:conf/eccv/WuCHLWSYL22,DBLP:conf/cvpr/YingMGB21}. As for VDPVE, we conducted ablation studies based on a custom train-test split, as the official validation and testing splits for the NTIRE 2023 VQA Challenge have not yet been released.
Particularly, we carefully divide the annotated 839 enhanced videos into two parts, 728 videos for training and 111 videos for testing. As shown in \cref{fig:mos,fig:methods}, the distributions of our train-test splits are consistent in MOS and video enhancement methods. 

\paragraph{Evaluation Criteria.} Spearman’s Rank-Order Correlation Coefficient (SRCC) and Pearson’s Linear Correlation Coefficient (PLCC) are selected as criteria to measure the accuracy and monotonicity, respectively. They are in the range of [0.0, 1.0]. A larger SRCC shows a more accurate ranking between different samples. A larger PLCC means a more accurate numerical fit with MOS scores. Besides, the mean average of PLCC and SRCC (named main score) is also reported as a comprehensive criterion.


\subsection{Implementation Details}

Our experiments are conducted on an NVIDIA V100 GPU with PyTorch 1.12 \cite{DBLP:conf/nips/PaszkeGMLBCKLGA19} and CUDA 11.3 for training and testing. 
All ablation studies are performed on the custom-divided train-test splits of VDPVE. 
As stated in \cref{sec:method}, the two branches of Zoom-VQA are trained independently and evaluated in a late-fusion manner. Below, we describe relevant settings in turn:

\vspace{-0.3cm}
\paragraph{IQA branch.} We choose the pre-trained ConvNext-Tiny from ImageNet-1k as the backbone. Following the standard training strategy \cite{DBLP:conf/cvpr/QPT, DBLP:conf/cvpr/YangWSLGCWY22}, we sample frames at 2 fps and resize the smaller edge of images to 512 while maintaining the aspect ratio. During training, images are cropped at a random size of $320\times320$ and subject to horizontal flipping with a probability of 0.5. We utilize the AdamW optimizer with a learning rate of 0.002, a weight decay of 0.01 and a cosine annealing scheduler. The mini-batch size is set to 32. During testing, the center-cropped patches of all extracted frames are used for evaluation.

\vspace{-0.3cm}
\paragraph{VQA branch.} We inherit most of the settings from FAST-VQA while enlarging the size of fragments and patch partition head. Specifically, we expand the path size from 4 to 6 and the corresponding fragment size from 32 to 48. During training, we choose the Video Swin Transformer-Tiny which is pre-trained on LSVQ. 
The AdamW optimizer is used for 30-epoch training, with a weight decay of 0.01 and a mini-batch size of 16. 
The initial global learning rate is 0.001 and decayed by a cosine annealing schedule. We sample a clip of 32 frames from each full-length video using a temporal stride of 2 and spatial size of $336 \times 336$. During testing, we infer 4 views for each video, and the final score is computed as the average score of all views. 

\begin{table*}[t]
    \centering
    \small
    \caption{Comparison with SOTA methods on public VQA benchmarks. LSVQ$_{\text{test}}$ and LSVQ$_{\text{1080p}}$ are official intra-dataset test subsets. While KoNViD-1k and LIVE-VQC are used for cross-dataset testing. The best and second best results are \textbf{bolded} and \underline{underlined}. Note that our reproduced results of FAST-VQA are a little lower than the official ones. We speculate that the reason for the difference is the variation in the positions of the sampled fragments (both in the spatial and temporal dimensions).}
    \begin{tabular}{c|cc|cc|cc|cc}
    \toprule
        Testing Type & \multicolumn{4}{c|}{Intra-dataset Test Sets} & \multicolumn{4}{c}{Cross-dataset Test Sets} \\
    \midrule
        Testing Set & \multicolumn{2}{c|}{LSVQ$_{\text{test}}$} & \multicolumn{2}{c|}
        {LSVQ$_{\text{1080p}}$} & \multicolumn{2}{c|}{KoNViD-1k} & \multicolumn{2}{c}{LIVE-VQC}\\
    \midrule
        Methods & SRCC & PLCC & SRCC & PLCC & SRCC & PLCC & SRCC & PLCC \\
    \midrule
        BRISQUE \cite{DBLP:journals/tip/MittalMB12} & 0.569 & 0.576 & 0.497 & 0.531 & 0.646 & 0.647 & 0.524 & 0.536 \\
        TLVQM \cite{DBLP:journals/tip/Korhonen19} & 0.772 & 0.774 & 0.589 & 0.616 & 0.732 & 0.724 & 0.670 & 0.691 \\
        VIDEVAL \cite{DBLP:journals/tip/TuWBAB21} & 0.794 & 0.783 & 0.545 & 0.554 & 0.751 & 0.741 & 0.630 & 0.640 \\
        VSFA \cite{DBLP:conf/mm/LiJJ19} & 0.801 & 0.796 & 0.675 & 0.704 & 0.784 & 0.794 & 0.734 & 0.772 \\
        PVQ$_{\text{wo/ patch}}$ \cite{DBLP:conf/cvpr/YingMGB21} & 0.814 & 0.816 & 0.686 & 0.708 & 0.781 & 0.781 & 0.747 & 0.776 \\
        PVQ$_{\text{w/ patch}}$ \cite{DBLP:conf/cvpr/YingMGB21} & 0.827 & 0.828 & 0.711 & 0.739 & 0.791 & 0.795 & 0.770 & 0.807 \\
        FAST-VQA-M \cite{DBLP:conf/eccv/WuCHLWSYL22} & 0.852 & 0.854 & 0.739 & 0.773 & 0.841 & 0.832 & 0.788 & 0.810 \\
        FAST-VQA \cite{DBLP:conf/eccv/WuCHLWSYL22} & \underline{0.876} & \underline{0.877} & 0.779 & \underline{0.814} & \underline{0.859} & \underline{0.855} & \textbf{0.823} & \textbf{0.844} \\
    \midrule
        FAST-VQA* (our reproduced) & 0.8686 & 0.8649 & 0.7707 & 0.7899 & 0.8418 & 0.8390 & 0.8124 & 0.8310 \\
        Zoom-VQA$_{w/ \text{IQA}}$ & 0.8583 & 0.8557 & 0.7469 & 0.7907 & 0.8436 & 0.8373 & 0.7675 & 0.7970 \\
        Zoom-VQA$_{w/ \text{VQA}}$ & 0.8735 & 0.8684 & \underline{0.7800} & 0.7965 & 0.8448 & 0.8452 & 0.8137 & 0.8319 \\
        \rowcolor{LightCyan} Zoom-VQA & \textbf{0.8860} & \textbf{0.8791} & \textbf{0.7985} & \textbf{0.8189} & \textbf{0.8772} & \textbf{0.8745} & \underline{0.8141} & \underline{0.8327} \\
    \bottomrule
    \end{tabular}
    \label{tab:sota}
    \vspace{-0.2cm}
\end{table*}

\subsection{Comparison with SOTA Results}

In \cref{tab:sota}, we compare with current SOTA methods, including classical and deep learning-based VQA methods. And 4 testing sets are used for evaluation models trained on LSVQ. Among them, LSVQ$_{\text{test}}$ and LSVQ$_{\text{1080p}}$ are official intra-dataset test subsets. While KoNViD-1k and LIVE-VQC are used for the evaluation of generalization ability through cross-dataset testing. Compared with classical methods that rely on statistical regularities (\eg, BRISQUE), Zoom-VQA outperforms by large margins. Compared with some deep learning-based methods that apply well-designed networks (\eg, TLVQM), Zoom-VQA still obtains higher performances. Compared with PVQ which also considers local and global perceptual qualities using 2D/3D features, Zoom-VQA obtains higher results, improving the SRCC from 0.827 to 0.886 (+0.059) in LSVQ$_{\text{test}}$. Furthermore, compared with the current SOTA FAST-VQA, \textbf{Zoom-VQA improves the best results from 0.876 to 0.886 of SRCC in LSVQ$_{\text{test}}$, 0.779 to 0.798 in LSVQ$_{\text{1080p}}$.} Even without fine-tuning, Zoom-VQA obtains 0.877 of SRCC in KoNViD-1k, which is higher than the current best results of FAST-VQA (0.859). These prove the generalization ability of Zoom-VQA.
Furthermore, we also report the results using separate IQA/VQA branches. The clip ensemble further boosts the final performance.

\subsection{Ablation Studies}

To further analyze the effectiveness of each component, we conduct ablation studies on the custom-divided train-test splits of VDPVE, which are illustrated in \cref{ssec:dataset}.

\begin{table}[t]
	\centering
	\small
	\caption{Ablation study on the clip ensemble in VDPVE. The proposed ensemble             strategy further boosts final performance.}
        \vspace{-0.2cm}
	\begin{tabular}{cc|cc|c}
		\toprule
            IQA & VQA & SRCC & PLCC & Main Score\\
            \midrule
            \checkmark & \checkmark & 0.7872 & 0.7897 & \textbf{0.7885} \\
            \midrule
            \checkmark & $\times$ & 0.7727 & 0.7867 & 0.7797 \\
            $\times$ & \checkmark & 0.7389 & 0.7449 & 0.7419 \\
		\bottomrule
	\end{tabular}
	\label{tab:ensemble}
\end{table}

\vspace{-0.3cm}
\paragraph{Ablation on the clip ensemble.} Both the results in \cref{tab:sota} and \cref{tab:ensemble} prove the effectiveness of the ensemble of the proposed IQA and VQA branches. By the proposed dual-branch design, Zoom-VQA can effectively capture both local and global information that impacts video quality.

\begin{table}[t]
	\centering
	\small
	\caption{
        Ablation study on the usage of patch attention module and feature pyramid alignment using the IQA branch in VDPVE.}
        \vspace{-0.2cm}
	\begin{tabular}{cc|cc|c}
		\toprule
            PAM & FPA & SRCC & PLCC & Main Score\\
            \midrule
            \checkmark & \checkmark & 0.7727 & 0.7867 & \textbf{0.7797}  \\
            \midrule
            \checkmark & $\times$   & 0.7576 & 0.7606 & 0.7591  \\
            $\times$ & \checkmark   & 0.7424 & 0.7438 & 0.7431  \\
            $\times$ & $\times$     & 0.7185 & 0.7438 & 0.7311  \\
		\bottomrule
	\end{tabular}
	\label{tab:pam}
        \vspace{-0.3cm}
\end{table}

\vspace{-0.3cm}
\paragraph{Ablation on the patch attention module and the feature pyramid alignment.} In \cref{tab:pam}, removing the patch head attention module (PAM) leads to a drop of 0.0366 (from 0.7797 to 0.7431) in the main score. It proves the importance of considering qualities in different spatial regions. While removing the feature pyramid alignment (FPA) module decreases the main score from 0.7797 to 0.7591, which verifies the utility of using multi-scale information for capturing distortions in different feature levels.


\begin{table}[t]
	\centering
	\small
	\caption{
            Ablation study on the partition head expansion in the VQA branch using different expansion types or window sizes.}
	\begin{tabular}{cc|cc|c}
		\toprule
            Expansion & Size & SRCC & PLCC & Main Score\\
            \midrule
            zero padding & 6 & 0.7389 & 0.7449 & \textbf{0.7419} \\
            zero padding & 8 & 0.7274 & 0.7352 & {0.7313} \\
            \midrule
            \textit{w/o} & 4 & 0.7244 & 0.7324 & 0.7284 \\
            reflect      & 6 & 0.6838 & 0.6709 & 0.6773 \\
            replicate    & 6 & 0.6973 & 0.7067 & 0.7020 \\
		\bottomrule
	\end{tabular}
	\label{tab:phe}
        \vspace{-0.2cm}
\end{table}

\paragraph{Ablation on the partition head expansion.} In \cref{tab:phe}, we test different types of expansion for the initialization of kernels in the linear embedding. And the used zero padding achieves the best performance. We claim that zero padding can maintain consistency in the distribution of features during the initialization stage, which is beneficial for subsequent network training. And further enlarging the window size does not introduce continuous improvement.

\subsection{NTIRE23 VQA Challenge Report}

The final ranking of the competition in the testing phase is shown in \cref{tab:final}. Our team QuoVadis, adopting Zoom-VQA, has achieved \textit{2nd place} based on the main score. This demonstrates the good generalization ability of Zoom-VQA when facing unseen domains.

\begin{table}[t]
	\centering
	\footnotesize
	\caption{Comparisons on the NTIRE23 VQA challenge. The results are given based on the testing phase. Zoom-VQA achieved 2nd place according to the main score.}
	\begin{tabular}{c|c|c}
		\toprule
		Team Name & Main Score & Ranking \\
		\midrule
            TB-VQA & 0.8576 & 1 \\ 
            \rowcolor{LightCyan} \textbf{QuoVadis (ours)} & 0.8396 & 2 \\
            OPDAI    & 0.8289 & 3 \\
            TIA Team & 0.8199 & 4 \\
            VCCIP   & 0.7994 & 5 \\
            IVL     & 0.7859 & 6 \\
            HXHHXH  & 0.7850 & 7 \\
            fmgtv	& 0.7727 & 8 \\
            KKARC	& 0.7635 & 9 \\
            DTVQA	& 0.7325 & 10 \\
            sqiyx	& 0.7302 & 11 \\
            402Lab	& 0.7136 & 12 \\
            one-for-all & 0.6990 & 13 \\
            NTU-SLab    & 0.6972 & 14 \\
            HNU-LIMMC	& 0.6923 & 15 \\
            Drealitym	& 0.6863 & 16 \\
            LION-Vaader	& 0.6596 & 17 \\
            Caption Timor & 0.6499 & 18 \\
            IVLab	      & 0.5851 & 19 \\
		\bottomrule
	\end{tabular}
	\label{tab:final}
        \vspace{-0.2cm}
\end{table}

\subsection{Visualization}
\label{sec:visul}

Following the visualization approach in \cite{DBLP:conf/eccv/WuCHLWSYL22}, we generate the spatial-temporal local quality map of our well-trained VQA branch. In this way, we can obtain an intuitive understanding of what is learned by Zoom-VQA. 

\begin{figure}[t]
  \centering
  \includegraphics[width=\linewidth]{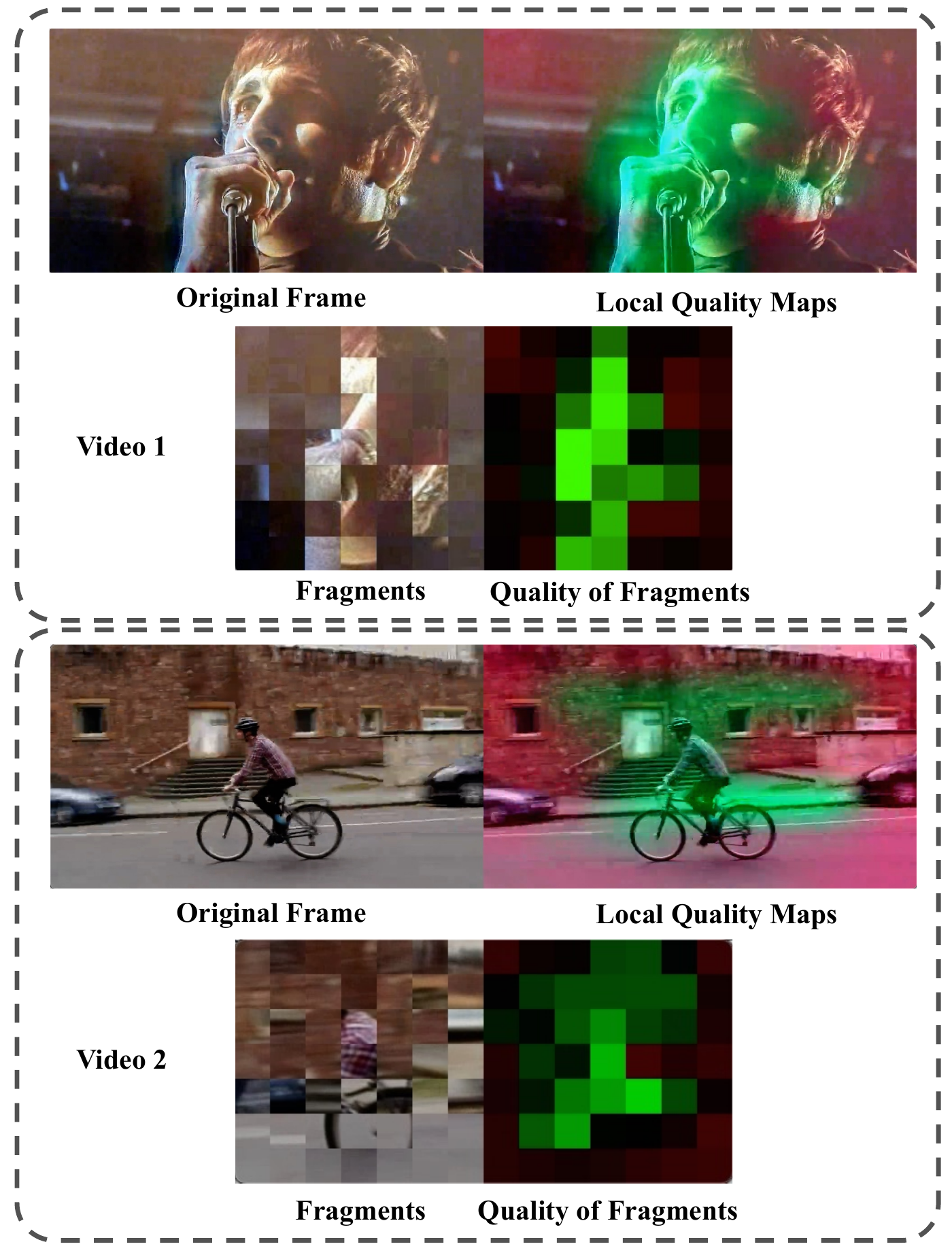}
  \caption{Spatio-temporal local quality maps, where \textcolor{red}{red} areas refer to relatively low quality scores and \textcolor{green}{green} areas refer to high scores. These demo videos are selected from VDPVE in 720P. }
  \label{fig:visualization}
  \vspace{-0.5cm}
\end{figure}

As shown in \cref{fig:visualization}, we sample three typical videos from VDPVE in 720p and plot the original frame, local quality map, fragments, and quality map of fragments column by column. From video 1, we can find our VQA model tends to concentrate on region-of-interest (\ie, face regions of the singer), which is consistent with HVS. Moreover, our VQA is assumed to be sensitive to textual quality information, since the qualities of clear areas are remarkably different from motion blur areas in video 2. 

\section{Conclusion}
\label{sec:conclusion}

In this paper, we proposed a novel Zoom-VQA framework that assesses video quality by comprehensively analyzing the ``patch-frame-clip" dimension of the video. Structurally, Zoom-VQA consists of two branches that receive image/clip input separately, considering both spatial and temporal information. Through designs of patch attention module, feature pyramid alignment, and clip ensemble, distortions in spatio-temporal and different feature levels can be effectively captured. Experimental results in VDPVE, LSVQ, KoNViD-1k, and LIVE-VQC prove the generalization ability of Zoom-VQA. Ablation studies further verify the effectiveness of each component.

\clearpage
{\small
\bibliographystyle{ieee_fullname}
\bibliography{egbib}

\begin{thebibliography}{10}\itemsep=-1pt

\bibitem{DBLP:conf/iccv/Arnab0H0LS21}
Anurag Arnab, Mostafa Dehghani, Georg Heigold, Chen Sun, Mario Lucic, and
  Cordelia Schmid.
\newblock Vivit: {A} video vision transformer.
\newblock In {\em {ICCV}}. {IEEE}, 2021.

\bibitem{DBLP:journals/tip/BosseMMWS18}
Sebastian Bosse, Dominique Maniry, Klaus{-}Robert M{\"{u}}ller, Thomas Wiegand,
  and Wojciech Samek.
\newblock Deep neural networks for no-reference and full-reference image
  quality assessment.
\newblock {\em {IEEE} TIP}, 27(1):206--219, 2018.

\bibitem{DBLP:conf/eccv/CarionMSUKZ20}
Nicolas Carion, Francisco Massa, Gabriel Synnaeve, Nicolas Usunier, Alexander
  Kirillov, and Sergey Zagoruyko.
\newblock End-to-end object detection with transformers.
\newblock In {\em {ECCV}}, pages 213--229. Springer, 2020.

\bibitem{DBLP:conf/cvpr/CarreiraZ17}
Jo{\~{a}}o Carreira and Andrew Zisserman.
\newblock Quo vadis, action recognition? {A} new model and the kinetics
  dataset.
\newblock In {\em {CVPR}}, pages 4724--4733. {IEEE} Computer Society, 2017.

\bibitem{DBLP:conf/cvpr/CheonYKL21}
Manri Cheon, Sung{-}Jun Yoon, Byungyeon Kang, and Junwoo Lee.
\newblock Perceptual image quality assessment with transformers.
\newblock In {\em {CVPR} Workshops}. {IEEE}, 2021.

\bibitem{DBLP:conf/emnlp/ChoMGBBSB14}
Kyunghyun Cho, Bart van Merrienboer, {\c{C}}aglar G{\"{u}}l{\c{c}}ehre, Dzmitry
  Bahdanau, Fethi Bougares, Holger Schwenk, and Yoshua Bengio.
\newblock Learning phrase representations using {RNN} encoder-decoder for
  statistical machine translation.
\newblock In {\em {EMNLP}}, pages 1724--1734. {ACL}, 2014.

\bibitem{DBLP:conf/cvpr/DengDSLL009}
Jia Deng, Wei Dong, Richard Socher, Li{-}Jia Li, Kai Li, and Li Fei{-}Fei.
\newblock Imagenet: {A} large-scale hierarchical image database.
\newblock In {\em {CVPR}}, pages 248--255. {IEEE}, 2009.

\bibitem{DBLP:conf/iclr/DosovitskiyB0WZ21}
Alexey Dosovitskiy, Lucas Beyer, Alexander Kolesnikov, Dirk Weissenborn,
  Xiaohua Zhai, Thomas Unterthiner, Mostafa Dehghani, Matthias Minderer, Georg
  Heigold, Sylvain Gelly, Jakob Uszkoreit, and Neil Houlsby.
\newblock An image is worth 16x16 words: Transformers for image recognition at
  scale.
\newblock In {\em {ICLR}}. OpenReview.net, 2021.

\bibitem{DBLP:conf/iccv/0001XMLYMF21}
Haoqi Fan, Bo Xiong, Karttikeya Mangalam, Yanghao Li, Zhicheng Yan, Jitendra
  Malik, and Christoph Feichtenhofer.
\newblock Multiscale vision transformers.
\newblock In {\em {ICCV}}, pages 6804--6815. {IEEE}, 2021.

\bibitem{DBLP:journals/corr/abs-1812-03982}
Christoph Feichtenhofer, Haoqi Fan, Jitendra Malik, and Kaiming He.
\newblock Slowfast networks for video recognition.
\newblock {\em CoRR}, abs/1812.03982, 2018.

\bibitem{gao2023vdpve}
Yixuan Gao, Yuqin Cao, Tengchuan Kou, Wei Sun, Yunlong Dong, Xiaohong Liu,
  Xiongkuo Min, and Guangtao Zhai.
\newblock Vdpve: Vqa dataset for perceptual video enhancement.
\newblock {\em CoRR}, abs/2303.09290, 2023.

\bibitem{DBLP:journals/corr/GhadiyaramB16}
Deepti Ghadiyaram and Alan~C. Bovik.
\newblock Perceptual quality prediction on authentically distorted images using
  a bag of features approach.
\newblock {\em CoRR}, abs/1609.04757, 2016.

\bibitem{DBLP:journals/access/Gotz-HahnHLS21}
Franz G{\"{o}}tz{-}Hahn, Vlad Hosu, Hanhe Lin, and Dietmar Saupe.
\newblock Konvid-150k: {A} dataset for no-reference video quality assessment of
  videos in-the-wild.
\newblock {\em {IEEE} Access}, 9:72139--72160, 2021.

\bibitem{DBLP:conf/cvpr/GuoBHZL21}
Haiyang Guo, Yi Bin, Yuqing Hou, Qing Zhang, and Hengliang Luo.
\newblock {IQMA} network: Image quality multi-scale assessment network.
\newblock In {\em {CVPR} Workshops}, pages 443--452. {IEEE}, 2021.

\bibitem{DBLP:conf/cvpr/HeSCFD22}
Jingwen He, Wu Shi, Kai Chen, Lean Fu, and Chao Dong.
\newblock {GCFSR:} a generative and controllable face super resolution method
  without facial and {GAN} priors.
\newblock In {\em {CVPR}}, pages 1879--1888. {IEEE}, 2022.

\bibitem{DBLP:conf/cvpr/HeZRS16}
Kaiming He, Xiangyu Zhang, Shaoqing Ren, and Jian Sun.
\newblock Deep residual learning for image recognition.
\newblock In {\em {CVPR}}, pages 770--778. {IEEE}, 2016.

\bibitem{DBLP:conf/qomex/HosuHJLMSLS17}
Vlad Hosu, Franz Hahn, Mohsen Jenadeleh, Hanhe Lin, Hui Men, Tam{\'{a}}s
  Szir{\'{a}}nyi, Shujun Li, and Dietmar Saupe.
\newblock The konstanz natural video database (konvid-1k).
\newblock In {\em QoMEX}, pages 1--6. {IEEE}, 2017.

\bibitem{DBLP:conf/iclr/KipfW17}
Thomas~N. Kipf and Max Welling.
\newblock Semi-supervised classification with graph convolutional networks.
\newblock In {\em {ICLR} (Poster)}. OpenReview.net, 2017.

\bibitem{DBLP:journals/tip/Korhonen19}
Jari Korhonen.
\newblock Two-level approach for no-reference consumer video quality
  assessment.
\newblock {\em {IEEE} Trans. Image Process.}, 28(12):5923--5938, 2019.

\bibitem{DBLP:conf/mm/KorhonenSY20}
Jari Korhonen, Yicheng Su, and Junyong You.
\newblock Blind natural video quality prediction via statistical temporal
  features and deep spatial features.
\newblock In {\em {ACM} MM}, 2020.

\bibitem{DBLP:journals/tcsv/LiZTZW22}
Bowen Li, Weixia Zhang, Meng Tian, Guangtao Zhai, and Xianpei Wang.
\newblock Blindly assess quality of in-the-wild videos via quality-aware
  pre-training and motion perception.
\newblock {\em {IEEE} Trans. Circ. Syst. Video Tech.}, 32(9):5944--5958, 2022.

\bibitem{DBLP:conf/mm/LiJJ19}
Dingquan Li, Tingting Jiang, and Ming Jiang.
\newblock Quality assessment of in-the-wild videos.
\newblock In {\em {ACM} Multimedia}, pages 2351--2359. {ACM}, 2019.

\bibitem{DBLP:journals/ijcv/LiJJ21}
Dingquan Li, Tingting Jiang, and Ming Jiang.
\newblock Unified quality assessment of in-the-wild videos with mixed datasets
  training.
\newblock {\em Int. J. Comput. Vis.}, 129(4):1238--1257, 2021.

\bibitem{DBLP:journals/ijon/LiYCCFXLC22}
Haoying Li, Yifan Yang, Meng Chang, Shiqi Chen, Huajun Feng, Zhihai Xu, Qi Li,
  and Yueting Chen.
\newblock Srdiff: Single image super-resolution with diffusion probabilistic
  models.
\newblock {\em Neurocomputing}, 479:47--59, 2022.

\bibitem{DBLP:conf/iccvw/LiangCSZGT21}
Jingyun Liang, Jiezhang Cao, Guolei Sun, Kai Zhang, Luc~Van Gool, and Radu
  Timofte.
\newblock Swinir: Image restoration using swin transformer.
\newblock In {\em {ICCV} Workshops}, pages 1833--1844. {IEEE}, 2021.

\bibitem{DBLP:conf/iccv/LiuL00W0LG21}
Ze Liu, Yutong Lin, Yue Cao, Han Hu, Yixuan Wei, Zheng Zhang, Stephen Lin, and
  Baining Guo.
\newblock Swin transformer: Hierarchical vision transformer using shifted
  windows.
\newblock In {\em {ICCV}}, pages 9992--10002. {IEEE}, 2021.

\bibitem{DBLP:conf/cvpr/0003MWFDX22}
Zhuang Liu, Hanzi Mao, Chao{-}Yuan Wu, Christoph Feichtenhofer, Trevor Darrell,
  and Saining Xie.
\newblock A convnet for the 2020s.
\newblock In {\em {CVPR}}, pages 11966--11976. {IEEE}, 2022.

\bibitem{DBLP:conf/cvpr/LiuN0W00022}
Ze Liu, Jia Ning, Yue Cao, Yixuan Wei, Zheng Zhang, Stephen Lin, and Han Hu.
\newblock Video swin transformer.
\newblock In {\em {CVPR}}, pages 3192--3201. {IEEE}, 2022.

\bibitem{DBLP:conf/mm/MaF21}
Kede Ma and Yuming Fang.
\newblock Image quality assessment in the modern age.
\newblock In {\em {ACM} MM}, pages 5664--5666, 2021.

\bibitem{DBLP:journals/tip/MadhusudanaBWAB22}
Pavan~C. Madhusudana, Neil Birkbeck, Yilin Wang, Balu Adsumilli, and Alan~C.
  Bovik.
\newblock Image quality assessment using contrastive learning.
\newblock {\em {IEEE} Trans. Image Process.}, 31:4149--4161, 2022.

\bibitem{DBLP:journals/tip/MittalMB12}
Anish Mittal, Anush~Krishna Moorthy, and Alan~Conrad Bovik.
\newblock No-reference image quality assessment in the spatial domain.
\newblock {\em {IEEE} TIP}, 21(12):4695--4708, 2012.

\bibitem{DBLP:journals/tip/MittalSB16}
Anish Mittal, Michele~A. Saad, and Alan~C. Bovik.
\newblock A completely blind video integrity oracle.
\newblock {\em {IEEE} Trans. Image Process.}, 25(1):289--300, 2016.

\bibitem{DBLP:journals/spl/MittalSB13}
Anish Mittal, Rajiv Soundararajan, and Alan~C. Bovik.
\newblock Making a "completely blind" image quality analyzer.
\newblock {\em {IEEE} Signal Process. Lett.}, 20(3):209--212, 2013.

\bibitem{DBLP:conf/nips/PaszkeGMLBCKLGA19}
Adam Paszke, Sam Gross, Francisco Massa, Adam Lerer, James Bradbury, Gregory
  Chanan, Trevor Killeen, Zeming Lin, Natalia Gimelshein, Luca Antiga, Alban
  Desmaison, Andreas K{\"{o}}pf, Edward~Z. Yang, Zachary DeVito, Martin Raison,
  Alykhan Tejani, Sasank Chilamkurthy, Benoit Steiner, Lu Fang, Junjie Bai, and
  Soumith Chintala.
\newblock Pytorch: An imperative style, high-performance deep learning library.
\newblock In {\em NeurIPS}, pages 8024--8035, 2019.

\bibitem{DBLP:conf/iccv/QiuYM17}
Zhaofan Qiu, Ting Yao, and Tao Mei.
\newblock Learning spatio-temporal representation with pseudo-3d residual
  networks.
\newblock In {\em {ICCV}}, pages 5534--5542. {IEEE}, 2017.

\bibitem{DBLP:journals/tip/SaadBC14}
Michele~A. Saad, Alan~C. Bovik, and Christophe Charrier.
\newblock Blind prediction of natural video quality.
\newblock {\em {IEEE} Trans. Image Process.}, 23(3):1352--1365, 2014.

\bibitem{DBLP:journals/tsp/SchusterP97}
Mike Schuster and Kuldip~K. Paliwal.
\newblock Bidirectional recurrent neural networks.
\newblock {\em {IEEE} Trans. Signal Process.}, 45(11):2673--2681, 1997.

\bibitem{DBLP:journals/tip/SinnoB19}
Zeina Sinno and Alan~Conrad Bovik.
\newblock Large-scale study of perceptual video quality.
\newblock {\em {IEEE} Trans. Image Process.}, 28(2):612--627, 2019.

\bibitem{DBLP:journals/tip/SinnoB19/live-vqc}
Zeina Sinno and Alan~Conrad Bovik.
\newblock Large-scale study of perceptual video quality.
\newblock {\em {IEEE} Trans. Image Process.}, 28(2):612--627, 2019.

\bibitem{DBLP:conf/mm/SunMLZ22}
Wei Sun, Xiongkuo Min, Wei Lu, and Guangtao Zhai.
\newblock A deep learning based no-reference quality assessment model for {UGC}
  videos.
\newblock In {\em {ACM} MM}, pages 856--865, 2022.

\bibitem{DBLP:conf/icml/TouvronCDMSJ21}
Hugo Touvron, Matthieu Cord, Matthijs Douze, Francisco Massa, Alexandre
  Sablayrolles, and Herv{\'{e}} J{\'{e}}gou.
\newblock Training data-efficient image transformers {\&} distillation through
  attention.
\newblock In {\em {ICML}}, pages 10347--10357. {PMLR}, 2021.

\bibitem{DBLP:conf/iccv/TranBFTP15}
Du Tran, Lubomir~D. Bourdev, Rob Fergus, Lorenzo Torresani, and Manohar Paluri.
\newblock Learning spatiotemporal features with 3d convolutional networks.
\newblock In {\em {ICCV}}, pages 4489--4497. {IEEE} Computer Society, 2015.

\bibitem{DBLP:conf/cvpr/TranWTRLP18}
Du Tran, Heng Wang, Lorenzo Torresani, Jamie Ray, Yann LeCun, and Manohar
  Paluri.
\newblock A closer look at spatiotemporal convolutions for action recognition.
\newblock In {\em {CVPR}}, pages 6450--6459. {IEEE}, 2018.

\bibitem{DBLP:conf/icip/TuCCBAB20}
Zhengzhong Tu, Chia{-}Ju Chen, Li{-}Heng Chen, Neil Birkbeck, Balu Adsumilli,
  and Alan~C. Bovik.
\newblock A comparative evaluation of temporal pooling methods for blind video
  quality assessment.
\newblock In {\em {ICIP}}, pages 141--145. {IEEE}, 2020.

\bibitem{DBLP:journals/tip/TuWBAB21}
Zhengzhong Tu, Yilin Wang, Neil Birkbeck, Balu Adsumilli, and Alan~C. Bovik.
\newblock {UGC-VQA:} benchmarking blind video quality assessment for user
  generated content.
\newblock {\em {IEEE} Trans. Image Process.}, 30:4449--4464, 2021.

\bibitem{DBLP:journals/corr/abs-2101-10955}
Zhengzhong Tu, Xiangxu Yu, Yilin Wang, Neil Birkbeck, Balu Adsumilli, and
  Alan~C. Bovik.
\newblock {RAPIQUE:} rapid and accurate video quality prediction of user
  generated content.
\newblock {\em CoRR}, abs/2101.10955, 2021.

\bibitem{DBLP:conf/iccvw/WangXDS21}
Xintao Wang, Liangbin Xie, Chao Dong, and Ying Shan.
\newblock Real-esrgan: Training real-world blind super-resolution with pure
  synthetic data.
\newblock In {\em {ICCV} Workshops}, pages 1905--1914. {IEEE}, 2021.

\bibitem{DBLP:conf/mmsp/WangIA19}
Yilin Wang, Sasi Inguva, and Balu Adsumilli.
\newblock Youtube {UGC} dataset for video compression research.
\newblock In {\em {MMSP}}, pages 1--5. {IEEE}, 2019.

\bibitem{DBLP:conf/eccv/WuCHLWSYL22}
Haoning Wu, Chaofeng Chen, Jingwen Hou, Liang Liao, Annan Wang, Wenxiu Sun,
  Qiong Yan, and Weisi Lin.
\newblock {FAST-VQA:} efficient end-to-end video quality assessment with
  fragment sampling.
\newblock In {\em {ECCV}}, pages 538--554. Springer, 2022.

\bibitem{DBLP:journals/corr/abs-2206-09853}
Haoning Wu, Chaofeng Chen, Liang Liao, Jingwen Hou, Wenxiu Sun, Qiong Yan, and
  Weisi Lin.
\newblock Discovqa: Temporal distortion-content transformers for video quality
  assessment.
\newblock {\em CoRR}, abs/2206.09853, 2022.

\bibitem{DBLP:journals/corr/abs-2211-04894}
Haoning Wu, Liang Liao, Chaofeng Chen, Jingwen Hou, Annan Wang, Wenxiu Sun,
  Qiong Yan, and Weisi Lin.
\newblock Disentangling aesthetic and technical effects for video quality
  assessment of user generated content.
\newblock {\em CoRR}, abs/2211.04894, 2022.

\bibitem{DBLP:journals/tog/XiaoKFCL18}
Lei Xiao, Anton Kaplanyan, Alexander Fix, Matthew Chapman, and Douglas Lanman.
\newblock Deepfocus: learned image synthesis for computational displays.
\newblock {\em {ACM} Trans. Graph.}, 37(6):200, 2018.

\bibitem{DBLP:journals/tog/XiaoNCFLK20}
Lei Xiao, Salah Nouri, Matthew Chapman, Alexander Fix, Douglas Lanman, and
  Anton Kaplanyan.
\newblock Neural supersampling for real-time rendering.
\newblock {\em {ACM} Trans. Graph.}, 39(4):142, 2020.

\bibitem{DBLP:conf/eccv/XieSHTM18}
Saining Xie, Chen Sun, Jonathan Huang, Zhuowen Tu, and Kevin Murphy.
\newblock Rethinking spatiotemporal feature learning: Speed-accuracy trade-offs
  in video classification.
\newblock In {\em {ECCV}}, pages 318--335. Springer, 2018.

\bibitem{DBLP:journals/corr/abs-2111-09886}
Zhenda Xie, Zheng Zhang, Yue Cao, Yutong Lin, Jianmin Bao, Zhuliang Yao, Qi
  Dai, and Han Hu.
\newblock Simmim: {A} simple framework for masked image modeling.
\newblock {\em CoRR}, abs/2111.09886, 2021.

\bibitem{DBLP:conf/mm/XuLZZW021}
Jiahua Xu, Jing Li, Xingguang Zhou, Wei Zhou, Baichao Wang, and Zhibo Chen.
\newblock Perceptual quality assessment of internet videos.
\newblock In {\em {ACM} MM}, pages 1248--1257, 2021.

\bibitem{DBLP:conf/cvpr/YangWSLGCWY22}
Sidi Yang, Tianhe Wu, Shuwei Shi, Shanshan Lao, Yuan Gong, Mingdeng Cao, Jiahao
  Wang, and Yujiu Yang.
\newblock {MANIQA:} multi-dimension attention network for no-reference image
  quality assessment.
\newblock In {\em {CVPR} Workshops}, pages 1190--1199. {IEEE}, 2022.

\bibitem{DBLP:conf/cvpr/YeKKD12}
Peng Ye, Jayant Kumar, Le Kang, and David~S. Doermann.
\newblock Unsupervised feature learning framework for no-reference image
  quality assessment.
\newblock In {\em {CVPR}}, pages 1098--1105. {IEEE} Computer Society, 2012.

\bibitem{DBLP:conf/cvpr/YingMGB21}
Zhenqiang Ying, Maniratnam Mandal, Deepti Ghadiyaram, and Alan~C. Bovik.
\newblock Patch-vq: 'patching up' the video quality problem.
\newblock In {\em {CVPR}}, pages 14019--14029. Computer Vision Foundation /
  {IEEE}, 2021.

\bibitem{DBLP:conf/iccv/YuanG0ZYW21}
Kun Yuan, Shaopeng Guo, Ziwei Liu, Aojun Zhou, Fengwei Yu, and Wei Wu.
\newblock Incorporating convolution designs into visual transformers.
\newblock In {\em {ICCV}}, pages 559--568. {IEEE}, 2021.

\bibitem{DBLP:conf/cvpr/QPT}
Kai Zhao, Kun Yuan, Ming Sun, Mading Li, and Xing Wen.
\newblock Quality-aware pre-trained models for blind image quality assessment.
\newblock In {\em {CVPR}}. {IEEE}, 2023.

\end{thebibliography}
}

\end{document}